\documentclass[12pt]{article}
\usepackage[T1]{fontenc}
\usepackage{lmodern}
\usepackage[margin=1in]{geometry}
\usepackage{setspace}
\usepackage{enumitem}
\usepackage{amsmath,amssymb,amsthm}
\usepackage{tikz}
\usetikzlibrary{arrows.meta,positioning}
\usepackage{sgame}
\usepackage[round,authoryear]{natbib}
\usepackage{xcolor}
\usepackage[colorlinks=true,citecolor=blue,linkcolor=red,urlcolor=blue]{hyperref}
\theoremstyle{definition}
\newtheorem{theorem}{Theorem}

\newcommand{\red}[1]{{\color{red}#1}}
\newcommand{\blue}[1]{{\color{blue}#1}}

\newcommand{\Prob}{\mathbb{P}}

\title{Neural networks and games}
\author{Philip R. Neary\footnote{Economics Department, Royal Holloway University of London}}
\date{\today}
\begin{document}
\maketitle
\begin{abstract}
\noindent
Hopfield neural networks and noncooperative games provide seemingly different approaches to understanding how collective outcomes emerge from local interactions.
In a neural network, individual units adjust their states according to local signals; in a game, players choose actions to maximise individual payoffs.
I establish an exact equivalence between symmetric binary Hopfield networks and binary-action games that admit a quadratic potential.
The equivalence identifies stable network configurations with pure-strategy Nash equilibria, and asynchronous neural updating with best-response dynamics.
More broadly, the storage and retrieval of memories in neural networks correspond, respectively, to the design and selection of equilibria in games.
\end{abstract}


\newpage
\section{Introduction}\label{sec:intro}

\citeauthor{Hopfield1982PNAS}'s (\citeyear{Hopfield1982PNAS}) foundational neural-network model
shows how simple, locally interacting units can
collectively store and retrieve information.
Neurons take one of two states and update in response
to their neighbours. These updates reduce a global
energy function, leading to stable configurations
that can serve as stored memories.

There is a natural parallel in noncooperative game theory, where players adjust their actions in response to one another.
In potential games \citep{MondererShapley1996GEB}, these adjustments can be captured by a single function that increases whenever a player makes a profitable deviation.
In finite games, any sequence of such improvements eventually lead to a Nash equilibrium \citep[Lemma~2.3]{MondererShapley1996GEB}.

In this note I show that, despite their different origins and terminology, these two approaches to decentralised interaction are mathematically equivalent.
Specifically, Theorem~\ref{thm:main} confirms that symmetric binary Hopfield networks are
equivalent to binary-action games whose incentives admit a quadratic potential.\footnote{Importantly, I have not found this equivalence explicitly stated in the literature.}
In particular, the negative of the Hopfield energy function is the game-theoretic potential, stable network configurations are Nash equilibria, and asynchronous neuronal updating corresponds to best-response dynamics.
More broadly, the storage and retrieval of memories in neural networks correspond, respectively, to the design of equilibria and their selection using dynamics.
My hope is that methods and questions developed in either literature may prove useful in the other.


\section{Neural nets as games and games as neural nets}\label{sec:equivalence}
Consider a finite set $V=\{1,\ldots,n\}$ of interacting units.
Each unit $i \in V$ is in one of two states, $x_i\in\{-1,+1\}$, and, as is standard, $\boldsymbol{x}_{-i}$ denotes the states of all units other than $i$.
The units are connected by an undirected graph $G=(V,E)$, where an edge $ij\in E$ indicates that units $i$ and $j$ interact and $N_i=\{j:ij\in E\}$ denotes the neighbours of $i$.

Now consider two interpretations of this environment.
In the first, the units are neurons whose states evolve in response to their neighbours.
In the second, they are players whose actions depend on the actions of those with whom they interact.
In the neural-network interpretation, behaviour is governed by an energy function defined over network configurations; in the game-theoretic variant, each player has a utility function that depends on her own action and those of her neighbours.

The connection between the two models can be seen clearly by presenting the ingredients in parallel.
The formal equivalence is established in Theorem~\ref{thm:main} that follows.

\par\medskip
\begingroup
\setlength{\columnsep}{0.04\textwidth}
\newcommand{\pairrow}[2]{%
  \noindent\begin{minipage}[t]{0.48\textwidth}\vspace{0pt}\raggedright\small\setstretch{1.12}#1\end{minipage}%
  \hfill\begin{minipage}[t]{0.48\textwidth}\vspace{0pt}\raggedright\small\setstretch{1.12}#2\end{minipage}\par\medskip}
\pairrow{\centering\textbf{Neural networks}}{\centering\textbf{Games}}
\pairrow{
  Neuron $i\in V$ has state $x_i\in\{-1,+1\}$.\\
  A pattern is a configuration $\boldsymbol{x}=(x_i)_{i\in V}$.
}{
  Player $i\in V$ chooses action $x_i\in\{-1,+1\}$.\\
  A pure strategy profile is $\boldsymbol{x}=(x_i)_{i\in V}$.
}
\pairrow{
  An edge $ij\in E$ represents a synaptic connection, with symmetric weight $J_{ij}=J_{ji}$. Set $J_{ij}=0$ off $E$ and $J_{ii}=0$.
}{
  An edge $ij\in E$ represents a bilateral game, with symmetric interaction coefficient $J_{ij}=J_{ji}$. Set $J_{ij}=0$ off $E$ and $J_{ii}=0$.
}
\pairrow{
  The local field of neuron $i$ is
  \[\ell_i(\boldsymbol{x}_{-i})=h_i+\sum_{j\in N_i}J_{ij}x_j.\]
}{
Player $i$ receives additive bilateral payoffs:
  \[u_i(\boldsymbol{x})=h_ix_i+\sum_{j\in N_i}J_{ij}x_ix_j.\]
}
\pairrow{
The neuron adopts the sign of its local field, adjusting its state accordingly; when the field is zero, it retains its state.
}{
  The player chooses a best response to her neighbours' actions; when indifferent, she retains her action.
}
\pairrow{
  The network has energy
  \[H(\boldsymbol{x})=-\sum_{ij\in E}J_{ij}x_ix_j-\sum_i h_ix_i.\]
}{
  The game has exact potential
  \[\Phi(\boldsymbol{x})=\sum_{ij\in E}J_{ij}x_ix_j+\sum_i h_ix_i.\]
}
\pairrow{
  A unilateral change of state gives
  \[\Delta_i H=-(x_i'-x_i)\ell_i(x_{-i}).\]
  Every strict asynchronous update strictly lowers energy.
}{
  A unilateral change of action gives
  \[\Delta_i u_i=\Delta_i\Phi=(x_i'-x_i)\ell_i(x_{-i}).\]
  Every strict best-response strictly raises potential.
}
\pairrow{
  A configuration is stable if no neuron would change when updated.
}{
  A profile is a Nash equilibrium if no player has a profitable unilateral deviation.
}
\endgroup

\bigskip

A player's payoff depends on her own action, whereas, interestingly, a neuron's local field is independent of its own state.
The two are related by $u_i(\boldsymbol{x})=x_i\ell_i(\boldsymbol{x}_{-i})$.

\begin{theorem}[Neural-network--game equivalence]\label{thm:main}
The following two systems are mathematically equivalent:
\begin{enumerate}
\item[(i)]\label{thm:H}
Symmetric binary Hopfield networks with $J_{ij}=J_{ji}$, $J_{ii}=0$, biases $h_i$, and energy
\[
H(\boldsymbol{x})=-\sum_{i<j}J_{ij}x_ix_j-\sum_i h_ix_i.
\]
\item[(ii)]\label{thm:phi}
 Binary-action exact-potential games admitting a quadratic potential\footnote{An exact potential is unique up to the addition of a constant that I set to zero for simplicity.}
\[
\Phi(\boldsymbol{x})=\sum_{i<j}J_{ij}x_ix_j+\sum_i h_ix_i.
\]
\end{enumerate}
\end{theorem}
\medskip

\begin{proof}
For any unilateral change from $x_i$ to $x_i'$,
\[
\Phi(x_i',\boldsymbol{x}_{-i})
-\Phi(x_i,\boldsymbol{x}_{-i})
=
(x_i'-x_i)
\left(h_i+\sum_{j\ne i}J_{ij}x_j\right)
=
-\Delta_i H.
\]
The middle expression is also the change in player
$i$'s payoff in the corresponding game.
Thus, the Hopfield network and the game have identical
unilateral incentives. Consequently, best responses
coincide with local-field updates, and it follows that stable
configurations coincide with Nash equilibria.
\end{proof}

To illustrate the equivalence, consider the threshold model of \citet{NearyNewton2017JMID}.
Along every edge $ij\in E$, neighbours play the following two-player game, where $\gamma_i,\gamma_j\in(0,1)$.
\begin{center}
\begin{game}{2}{2}[Player $i$][Player $j$]
 & $+1$ & $-1$\\
$+1$ & $1-\gamma_i,\;1-\gamma_j$ & $0,0$\\
$-1$ & $0,0$ & $\gamma_i,\;\gamma_j$
\end{game}
\end{center}
\medskip
Each player must use the same action with each neighbour and receives the sum of her bilateral payoffs. Writing $d_i$ for player $i$'s degree, the payoff difference is
\[
u_i(+1,\boldsymbol{x}_{-i})-u_i(-1,\boldsymbol{x}_{-i})
=\frac12\left[\sum_{j\in N_i}x_j+d_i(1-2\gamma_i)\right].
\]
Comparing this with the payoff difference implied by
the corresponding Hopfield network,
\[
2\left(h_i+\sum_{j\in N_i}J_{ij}x_j\right)
=
\frac12\left[
\sum_{j\in N_i}x_j+d_i(1-2\gamma_i)
\right],
\]
we get that $J_{ij}=1/4$ for every edge $ij$ and $h_i=d_i(1-2\gamma_i)/4$ for every vertex $i$.


\section{Associative memory and mechanism design}\label{sec:memory}

One of the principal applications of Hopfield networks is associative memory:~the ability to store patterns of neuronal states and subsequently retrieve them from incomplete or corrupted inputs. This raises three related questions: how to design a network that stores prescribed memories, how to account for unintended stable configurations, and how the network's dynamics determine which memory is retrieved.
To a game theorist, the equivalence established in Section~\ref{sec:equivalence} naturally brings to mind mechanism design, unintended equilibria, and equilibrium selection, respectively.

\subsection{Storing memories is a design problem}\label{ssec:storing}

Given a prescribed collection of patterns, the network designer seeks to choose the weights and
biases so that each pattern is stable.
By Theorem~\ref{thm:main}, this is equivalent to designing a game by choosing its interaction coefficients and individual payoff parameters so that the corresponding pure strategy profiles are Nash equilibria.
Memory storage is therefore simply, in game-theoretic jargon, a mechanism design problem.

Formally, let $\mathcal{M}\subseteq\{-1,+1\}^n$
denote the collection of prescribed memories the designer
wishes to store.
Writing $\boldsymbol{J}=(J_{ij})$ and $\boldsymbol{h}=(h_i)$ for a network's weight matrix and bias vector, the designer's problem is
\[
\begin{aligned}
\text{find}\quad &
\boldsymbol{J},\ \boldsymbol{h}\\
\text{subject to}\quad &
x_i\left(\sum_{j\neq i}J_{ij}x_j+h_i\right)\geq 0,
\quad
\text{for all }i\in V,\;
\boldsymbol{x}\in\mathcal{M},\\
&
J_{ij}=J_{ji},\qquad J_{ii}=0.
\end{aligned}
\]
Equivalently, let
$\operatorname{NE}(\boldsymbol{J},\boldsymbol{h})$
denote the set of pure Nash equilibria of the
corresponding game. The designer seeks
$\boldsymbol{J}$ and $\boldsymbol{h}$ such that
\[
\mathcal{M}\subseteq
\operatorname{NE}(\boldsymbol{J},\boldsymbol{h}).
\]

There is, however, an unusual feature of this mechanism design problem worth highlighting.
\citet{MaskinSjostrom2002HSCW} is a survey on implementation theory that opens as follows: ``The implementation problem is the problem of designing a mechanism (game form) such that the equilibrium outcomes satisfy a criterion of social optimality embodied in a social choice rule.''
In Hopfield networks the prescribed outcomes are simply the patterns the designer wishes to store.
But, in contrast, there is no requirement that these outcomes be socially optimal, or even that their welfare be worthy of consideration.
All that is demanded is equilibrium.

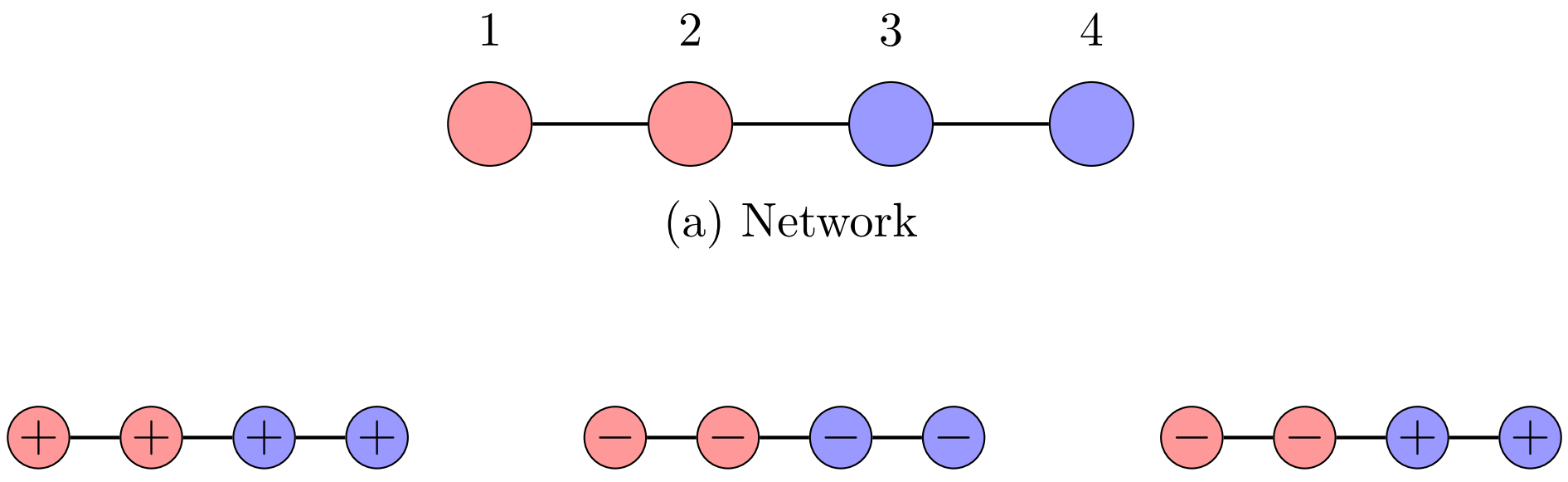
\begin{figure}[hbt!]
\centering

\begin{tikzpicture}[
    memoryH/.style={
        circle, draw,
        top color=red!40,
        bottom color=red!40,
        minimum size=14pt,
        inner sep=0pt
    },
    memoryL/.style={
        circle, draw,
        top color=blue!40,
        bottom color=blue!40,
        minimum size=14pt,
        inner sep=0pt
    },
    memoryTop/.style={
        minimum size=19pt
    }
]


\begin{scope}[shift={(3.6,2.5)}]
\node (a1) at (0,0) [memoryH,memoryTop] {};
\node (a2) at (1.6,0) [memoryH,memoryTop] {};
\node (a3) at (3.2,0) [memoryL,memoryTop] {};
\node (a4) at (4.8,0) [memoryL,memoryTop] {};

\draw[thick] (a1)--(a2)--(a3)--(a4);

\node[above=4pt] at (a1.north) {\small $1$};
\node[above=4pt] at (a2.north) {\small $2$};
\node[above=4pt] at (a3.north) {\small $3$};
\node[above=4pt] at (a4.north) {\small $4$};

\node at (2.4,-0.8) {\small (a) Network};
\end{scope}


\begin{scope}[shift={(0,0)}]
\node (b1) at (0,0) [memoryH] {$+$};
\node (b2) at (0.9,0) [memoryH] {$+$};
\node (b3) at (1.8,0) [memoryL] {$+$};
\node (b4) at (2.7,0) [memoryL] {$+$};

\draw[thick] (b1)--(b2)--(b3)--(b4);

\node at (1.35,-0.8)
    {\small (b) $\boldsymbol{x}^+$};
\end{scope}


\begin{scope}[shift={(4.6,0)}]
\node (c1) at (0,0) [memoryH] {$-$};
\node (c2) at (0.9,0) [memoryH] {$-$};
\node (c3) at (1.8,0) [memoryL] {$-$};
\node (c4) at (2.7,0) [memoryL] {$-$};

\draw[thick] (c1)--(c2)--(c3)--(c4);

\node at (1.35,-0.8)
    {\small (c) $\boldsymbol{x}^-$};
\end{scope}


\begin{scope}[shift={(9.2,0)}]
\node (d1) at (0,0) [memoryH] {$-$};
\node (d2) at (0.9,0) [memoryH] {$-$};
\node (d3) at (1.8,0) [memoryL] {$+$};
\node (d4) at (2.7,0) [memoryL] {$+$};

\draw[thick] (d1)--(d2)--(d3)--(d4);

\node at (1.35,-0.8)
    {\small (d) $\boldsymbol{x}^*$};
\end{scope}

\end{tikzpicture}

\caption{Memory storage on a four-player path.
Red vertices have threshold $3/4$, while blue vertices
have threshold $1/4$. Panel (a) shows the
network, with player indices above the vertices.
Panels (b) and (c) show the two prescribed memories,
while panel (d) shows an unintended stable configuration.
The signs inside the vertices indicate each neuron's state.}
\label{fig:memory}
\end{figure}

As an example, suppose a designer wishes to store the two patterns $\boldsymbol{x}^+=(+1,+1,+1,+1)$ and $\boldsymbol{x}^-=(-1,-1,-1,-1)$.
It is easy to see that many choices of weights and biases make both prescribed patterns stable.\footnote{If weak stability is permitted, the problem is trivial: setting all weights and biases to zero makes every configuration stable. Requiring strict stability instead gives a linear programming feasibility problem, which need not have a solution.}
As one possible solution, suppose the designer
chooses weights and biases corresponding to the
four-player threshold game in Figure~\ref{fig:memory},
with $\gamma_1=\gamma_2=3/4$ and
$\gamma_3=\gamma_4=1/4$.
Players 1 and 2 are coloured \red{red}, while players
3 and 4 are coloured \blue{blue}.\footnote{With only two values of $\gamma$, each either side of $\frac{1}{2}$, this is just the language game of \citep{Neary2011PhD,Neary2012GEB}.}
Both prescribed patterns, $\boldsymbol{x}^+$ and $\boldsymbol{x}^-$, are strict Nash equilibria of this game and hence stable configurations of the Hopfield network.

The example illustrates the distinction between choosing a game and choosing an equilibrium.
The designer chooses weights and biases which build a game, which in turn determine the equilibrium set.
The potential structure guarantees convergence to an equilibrium (with probability one) under 
asynchronous best-response dynamics.
But two issues remain.
First, satisfying the designer's constraints does not rule out additional
equilibria: the inclusion $\mathcal{M}\subseteq \operatorname{NE}(\boldsymbol{J},\boldsymbol{h})$ may be strict. 
Second, whether or not the inclusion is strict, when multiple equilibria exist, the equilibrium reached may depend on the initial configuration and the updating sequence.

Both issues arise in the example.
The resulting game admits a third equilibrium, $\boldsymbol{x}^*=(-1,-1,+1,+1)$, which was not among the prescribed memories.
Moreover, $\boldsymbol{x}^*$ is the unique potential maximiser.
Potential maximisation is a standard equilibrium-selection criterion and yet here it selects an outcome the designer did not intend.
I now turn to such unintended stable configurations, known as \emph{spurious memories} in the neural nets literature.

\subsection{Spurious memories are unintended equilibria}\label{ssec:unintended}

The problem with spurious memories is not simply that they exist, but that they can interfere with retrieval.
In the four-player example, the designer only wanted to store two prescribed memories, yet the resulting game admits three Nash equilibria.
This is why spurious memories can be problematic:~when presented with an incomplete or corrupted version of a prescribed memory, the network may instead retrieve the spurious memory.

This suggests a stronger version of the memory-storage problem is needed.
That is, rather than merely requiring that every prescribed pattern be stable, the designer may wish to ensure that these are the only stable configurations.
The mechanism design problem then becomes to construct a game wherein $\mathcal{M} = \operatorname{NE}(\boldsymbol{J},\boldsymbol{h})$.

\subsection{Retrieving memories and equilibrium selection}\label{ssec:selection}

Once a network has been designed, the problem of memory retrieval can be interpreted as a kind of \emph{targeted equilibrium selection}.
``Equilibrium'' because, by definition, any stored memory is a stable configuration, and ``targeted selection'' because the objective is to recover a particular memory identified by the input supplied to the network.

The way in which this differs from the quintessential equilibrium-selection problem in game theory is twofold.
First, the target may change from one moment (``retrieval episode'') to the next, even though the underlying game remains fixed.
Second, and more interestingly in my view, successful retrieval need not involve exact convergence to the targeted equilibrium:~recovering a close approximation, with high probability, is often deemed enough.\footnote{This leaves open how close ought be close enough. A question I will come back to.}

Suppose a network has been designed to store images
of a cat, a dog, and a horse. (See Figure~\ref{fig:animalmemories} in the Appendix.)
When presented with a corrupted image of a cat, we want the network to retrieve the cat; when subsequently presented with a corrupted image of a dog, we want it to retrieve the dog; etc.
Each retrieval begins with a new externally supplied input, rather than the configuration reached after the previous retrieval.
The justification for why state nearby the target is deemed ``good enough'' follows fairly naturally: since the memories are images represented by pixels, recovering an image with a small number of incorrect pixels may be perfectly adequate for recognition.

Let's return to the four-player example.
Suppose the designer wishes to retrieve the prescribed memory $\boldsymbol{x}^-=(-1,-1,-1,-1)$, and suppose further that the initial configuration is $\boldsymbol{y}=(-1,-1,-1,+1)$ which differs in the third coordinate.
Under asynchronous best-response dynamics, retrieval is not guaranteed.
If player 4 revises before player 3, the dynamics converge to $\boldsymbol{x}^-$ as intended.
But if player 3 revises first then the system converges to the spurious memory $\boldsymbol{x}^*=(-1,-1,+1,+1)$.
Each outcome occurs with probability $1/2$.

\begin{figure}[hbt!]
\centering

\begin{tikzpicture}[
    memoryH/.style={
        circle, draw,
        top color=red!40,
        bottom color=red!40,
        minimum size=15pt,
        inner sep=0pt
    },
    memoryL/.style={
        circle, draw,
        top color=blue!40,
        bottom color=blue!40,
        minimum size=15pt,
        inner sep=0pt
    }
]


\begin{scope}[shift={(3.4,2.8)}]
\node (a1) at (0,0) [memoryH] {$-$};
\node (a2) at (0.9,0) [memoryH] {$-$};
\node (a3) at (1.8,0) [memoryL] {$-$};
\node (a4) at (2.7,0) [memoryL] {$+$};

\draw[thick] (a1)--(a2)--(a3)--(a4);

\node at (1.35,0.75)
    {\small Initial configuration $\boldsymbol{y}$};
\end{scope}


\begin{scope}[shift={(0,0)}]
\node (b1) at (0,0) [memoryH] {$-$};
\node (b2) at (0.9,0) [memoryH] {$-$};
\node (b3) at (1.8,0) [memoryL] {$-$};
\node (b4) at (2.7,0) [memoryL] {$-$};

\draw[thick] (b1)--(b2)--(b3)--(b4);

\node at (1.35,-0.75)
    {\small Target memory $\boldsymbol{x}^-$};
\end{scope}


\begin{scope}[shift={(6.8,0)}]
\node (c1) at (0,0) [memoryH] {$-$};
\node (c2) at (0.9,0) [memoryH] {$-$};
\node (c3) at (1.8,0) [memoryL] {$+$};
\node (c4) at (2.7,0) [memoryL] {$+$};

\draw[thick] (c1)--(c2)--(c3)--(c4);

\node at (1.35,-0.75)
    {\small Spurious memory $\boldsymbol{x}^*$};
\end{scope}


\draw[->,thick]
    (4.2,2.35) -- (1.9,0.55)
    node[midway,left]
    {\small Player 4 first};

\draw[->,thick]
    (5.3,2.35) -- (8.2,0.55)
    node[midway,right]
    {\small Player 3 first};

\node at (1.35,0.85) {\small $\Prob=1/2$};
\node at (8.75,0.85) {\small $\Prob=1/2$};

\end{tikzpicture}

\caption{Targeted equilibrium selection in the
four-player example. The initial configuration is
at Hamming distance one from the target memory
$\boldsymbol{x}^-$. Under uniform random asynchronous
best-response dynamics, the target memory is retrieved
with probability $1/2$, while the spurious memory
$\boldsymbol{x}^*$ is selected with probability $1/2$.}
\label{fig:retrieval}
\end{figure}
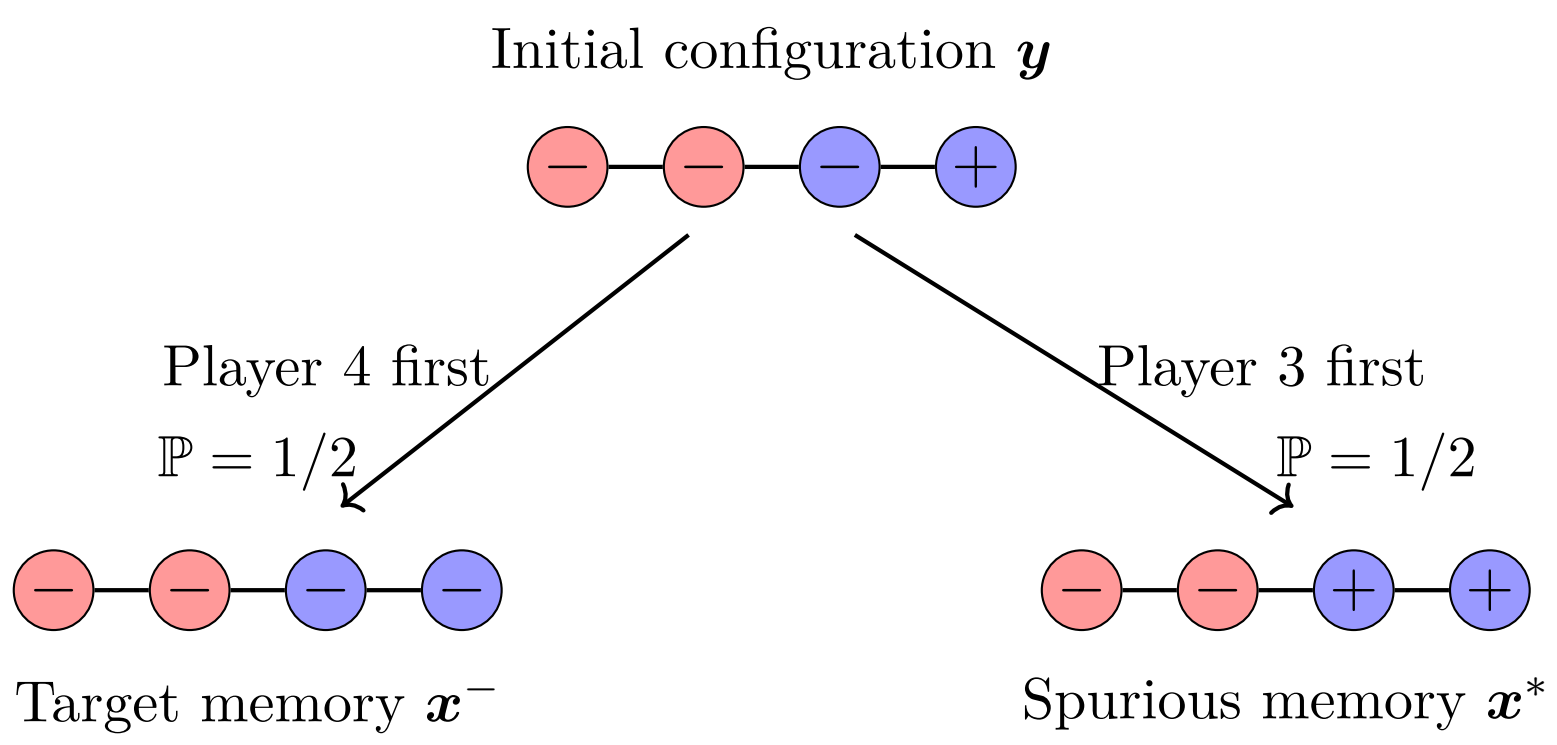

More generally, for a prescribed memory
$\boldsymbol{x}^\mu$ and an initial configuration
$\boldsymbol{y}$, define
\[
q_\mu(\boldsymbol{y})
=
\Prob\left(
\boldsymbol{X}_\infty=\boldsymbol{x}^\mu
\mid \boldsymbol{X}_0=\boldsymbol{y}
\right),
\]
where $\boldsymbol{X}_t$ follows random asynchronous
best-response dynamics. This is the probability of
successfully retrieving the targeted memory from
$\boldsymbol{y}$.

The example illustrates that Hamming distance and
retrieval probability capture different aspects of
the problem.\footnote{The Hamming distance between two configurations
$\boldsymbol{x},\boldsymbol{y}\in\{-1,+1\}^n$ is
$d_H(\boldsymbol{x},\boldsymbol{y})
=|\{i\in V:x_i\neq y_i\}|$.}
Hamming distance measures the extent
of corruption of the initial configuration, whereas
$q_\mu(\boldsymbol{y})$ measures the probability
that the dynamics recover the intended memory.
In the example, a configuration at Hamming distance one from the target only retrieved it with probability only $1/2$.
Indeed, it is easy to construct examples for which the initial configuration is at Hamming distance one from a strict equilibrium, yet the probability of retrieving that equilibrium is arbitrarily close to zero.


\section{So What?}

What do we learn from all this?
The mathematical equivalence established in Theorem~\ref{thm:main} is straightforward, and so it is not unreasonable to ask what is gained by making it explicit.
One immediate consequence is that we now have a common mathematical language connecting two previously distinct literatures, allowing results from one to be translated directed into the other.
But translating a known result from one language into another does not make it new, nor does it necessarily make it interesting.
The potential value of the connection may, however, extend beyond such translations.

One interesting possibility arises from the fact that the two literatures have developed largely independently, motivated by different questions.
Game theorists care about incentives, the rules of interaction, and equilibrium, while researchers studying associative memory are concerned with storage, retrieval, and the behaviour of neural networks.
Yet, as we have seen, the two literatures study essentially the same mathematical objects, albeit from different perspectives.
The potential benefit is therefore not simply that existing answers can be translated, but that new lines of enquiry may emerge from bringing the two perspectives together.
Indeed, the most interesting consequences of the connection may be ones that are difficult to anticipate from within either literature alone.

\clearpage
\appendix
\section{Cat, dog, and horse images}\label{app:pics}

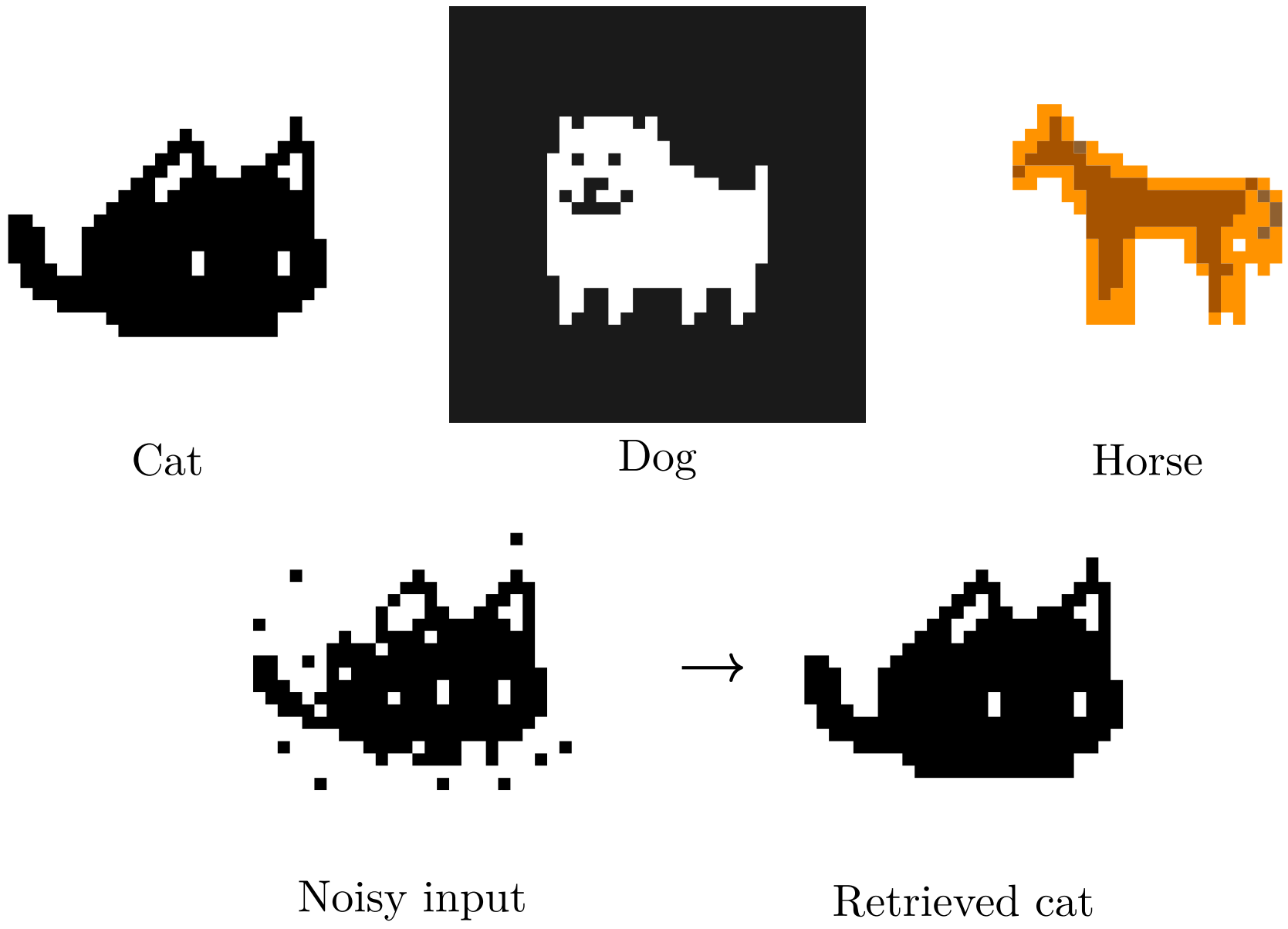
\begin{figure}[hbt!]
\centering
\begin{tikzpicture}[x=0.105cm,y=0.105cm]
\begin{scope}[shift={(3,41)}]
\fill[black]
  (24,19) rectangle (25,20)
  (15,18) rectangle (16,19)
  (24,18) rectangle (25,19)
  (14,17) rectangle (17,18)
  (23,17) rectangle (26,18)
  (13,16) rectangle (15,17)
  (16,16) rectangle (17,17)
  (22,16) rectangle (24,17)
  (25,16) rectangle (26,17)
  (12,15) rectangle (14,16)
  (16,15) rectangle (18,16)
  (20,15) rectangle (23,16)
  (25,15) rectangle (26,16)
  (11,14) rectangle (13,15)
  (15,14) rectangle (24,15)
  (25,14) rectangle (26,15)
  (10,13) rectangle (13,14)
  (14,13) rectangle (26,14)
  (9,12) rectangle (26,13)
  (1,11) rectangle (3,12)
  (8,11) rectangle (26,12)
  (1,10) rectangle (4,11)
  (7,10) rectangle (26,11)
  (1,9) rectangle (4,10)
  (7,9) rectangle (27,10)
  (1,8) rectangle (4,9)
  (7,8) rectangle (16,9)
  (17,8) rectangle (23,9)
  (24,8) rectangle (27,9)
  (2,7) rectangle (5,8)
  (7,7) rectangle (16,8)
  (17,7) rectangle (23,8)
  (24,7) rectangle (27,8)
  (2,6) rectangle (27,7)
  (3,5) rectangle (26,6)
  (5,4) rectangle (25,5)
  (9,3) rectangle (23,4)
  (10,2) rectangle (23,3)
;
\end{scope}
\node[font=\small] at (17,33) {Cat};
\begin{scope}[shift={(43,39)}]
\fill[black!90] (-3,-3) rectangle (31,31);
\fill[white]
  (6,21) rectangle (7,22)
  (8,21) rectangle (12,22)
  (13,21) rectangle (14,22)
  (6,20) rectangle (14,21)
  (6,19) rectangle (15,20)
  (5,18) rectangle (7,19)
  (8,18) rectangle (10,19)
  (11,18) rectangle (15,19)
  (5,17) rectangle (17,18)
  (22,17) rectangle (23,18)
  (5,16) rectangle (8,17)
  (10,16) rectangle (19,17)
  (22,16) rectangle (23,17)
  (5,15) rectangle (6,16)
  (7,15) rectangle (8,16)
  (9,15) rectangle (11,16)
  (12,15) rectangle (23,16)
  (5,14) rectangle (7,15)
  (11,14) rectangle (23,15)
  (5,13) rectangle (23,14)
  (5,12) rectangle (23,13)
  (5,11) rectangle (23,12)
  (5,10) rectangle (23,11)
  (5,9) rectangle (22,10)
  (6,8) rectangle (22,9)
  (6,7) rectangle (8,8)
  (10,7) rectangle (12,8)
  (16,7) rectangle (18,8)
  (20,7) rectangle (22,8)
  (6,6) rectangle (8,7)
  (10,6) rectangle (12,7)
  (16,6) rectangle (18,7)
  (20,6) rectangle (22,7)
  (6,5) rectangle (7,6)
  (10,5) rectangle (11,6)
  (16,5) rectangle (17,6)
  (20,5) rectangle (21,6)
;
\end{scope}
\node[font=\small] at (57,33) {Dog};
\begin{scope}[shift={(80,36)}]
\fill[orange!85!yellow]
  (8,25) rectangle (10,26)
  (8,24) rectangle (9,25)
  (10,24) rectangle (11,25)
  (7,23) rectangle (9,24)
  (10,23) rectangle (11,24)
  (6,22) rectangle (8,23)
  (12,22) rectangle (13,23)
  (6,21) rectangle (7,22)
  (12,21) rectangle (15,22)
  (7,20) rectangle (11,21)
  (14,20) rectangle (17,21)
  (6,19) rectangle (8,20)
  (10,19) rectangle (11,20)
  (17,19) rectangle (25,20)
  (26,19) rectangle (27,20)
  (10,18) rectangle (12,19)
  (25,18) rectangle (26,19)
  (27,18) rectangle (28,19)
  (11,17) rectangle (12,18)
  (25,17) rectangle (27,18)
  (24,16) rectangle (27,17)
  (16,15) rectangle (21,16)
  (23,15) rectangle (26,16)
  (27,15) rectangle (28,16)
  (12,14) rectangle (13,15)
  (15,14) rectangle (16,15)
  (20,14) rectangle (21,15)
  (23,14) rectangle (24,15)
  (25,14) rectangle (28,15)
  (12,13) rectangle (13,14)
  (15,13) rectangle (16,14)
  (20,13) rectangle (21,14)
  (23,13) rectangle (28,14)
  (12,12) rectangle (13,13)
  (15,12) rectangle (16,13)
  (21,12) rectangle (22,13)
  (24,12) rectangle (25,13)
  (26,12) rectangle (27,13)
  (12,11) rectangle (13,12)
  (15,11) rectangle (16,12)
  (23,11) rectangle (25,12)
  (12,10) rectangle (13,11)
  (14,10) rectangle (16,11)
  (23,10) rectangle (25,11)
  (12,9) rectangle (16,10)
  (23,9) rectangle (25,10)
  (12,8) rectangle (16,9)
  (22,8) rectangle (23,9)
  (24,8) rectangle (25,9)
;
\fill[orange!65!black]
  (9,24) rectangle (10,25)
  (9,23) rectangle (10,24)
  (8,22) rectangle (11,23)
  (7,21) rectangle (12,22)
  (6,20) rectangle (7,21)
  (11,20) rectangle (14,21)
  (11,19) rectangle (17,20)
  (25,19) rectangle (26,20)
  (12,18) rectangle (25,19)
  (12,17) rectangle (25,18)
  (12,16) rectangle (24,17)
  (12,15) rectangle (16,16)
  (21,15) rectangle (23,16)
  (13,14) rectangle (15,15)
  (21,14) rectangle (23,15)
  (13,13) rectangle (15,14)
  (21,13) rectangle (23,14)
  (13,12) rectangle (15,13)
  (22,12) rectangle (24,13)
  (13,11) rectangle (15,12)
  (22,11) rectangle (23,12)
  (13,10) rectangle (14,11)
  (22,10) rectangle (23,11)
  (22,9) rectangle (23,10)
;
\fill[brown!75!black]
  (11,22) rectangle (12,23)
  (26,18) rectangle (27,19)
  (27,17) rectangle (28,18)
  (27,16) rectangle (28,17)
  (26,15) rectangle (27,16)
;
\end{scope}
\node[font=\small] at (97,33) {Horse};
\begin{scope}[shift={(23,5)}]
\fill[black]
  (22,21) rectangle (23,22)
  (4,18) rectangle (5,19)
  (14,18) rectangle (15,19)
  (22,18) rectangle (23,19)
  (13,17) rectangle (16,18)
  (21,17) rectangle (24,18)
  (12,16) rectangle (13,17)
  (15,16) rectangle (16,17)
  (20,16) rectangle (22,17)
  (23,16) rectangle (24,17)
  (11,15) rectangle (12,16)
  (15,15) rectangle (17,16)
  (19,15) rectangle (21,16)
  (23,15) rectangle (24,16)
  (1,14) rectangle (2,15)
  (11,14) rectangle (12,15)
  (14,14) rectangle (22,15)
  (23,14) rectangle (24,15)
  (8,13) rectangle (9,14)
  (11,13) rectangle (15,14)
  (16,13) rectangle (24,14)
  (7,12) rectangle (11,13)
  (12,12) rectangle (24,13)
  (1,11) rectangle (3,12)
  (5,11) rectangle (6,12)
  (7,11) rectangle (24,12)
  (1,10) rectangle (3,11)
  (7,10) rectangle (8,11)
  (9,10) rectangle (25,11)
  (1,9) rectangle (4,10)
  (7,9) rectangle (16,10)
  (17,9) rectangle (21,10)
  (22,9) rectangle (25,10)
  (2,8) rectangle (5,9)
  (6,8) rectangle (12,9)
  (13,8) rectangle (16,9)
  (17,8) rectangle (21,9)
  (22,8) rectangle (25,9)
  (3,7) rectangle (6,8)
  (7,7) rectangle (25,8)
  (5,6) rectangle (24,7)
  (8,5) rectangle (23,6)
  (3,4) rectangle (4,5)
  (10,4) rectangle (14,5)
  (15,4) rectangle (18,5)
  (20,4) rectangle (21,5)
  (26,4) rectangle (27,5)
  (11,3) rectangle (12,4)
  (14,3) rectangle (18,4)
  (20,3) rectangle (21,4)
  (24,3) rectangle (25,4)
  (6,1) rectangle (7,2)
  (16,1) rectangle (17,2)
  (21,1) rectangle (22,2)
;
\end{scope}
\node[font=\small] at (37,-3) {Noisy input};
\begin{scope}[shift={(68,5)}]
\fill[black]
  (24,19) rectangle (25,20)
  (15,18) rectangle (16,19)
  (24,18) rectangle (25,19)
  (14,17) rectangle (17,18)
  (23,17) rectangle (26,18)
  (13,16) rectangle (15,17)
  (16,16) rectangle (17,17)
  (22,16) rectangle (24,17)
  (25,16) rectangle (26,17)
  (12,15) rectangle (14,16)
  (16,15) rectangle (18,16)
  (20,15) rectangle (23,16)
  (25,15) rectangle (26,16)
  (11,14) rectangle (13,15)
  (15,14) rectangle (24,15)
  (25,14) rectangle (26,15)
  (10,13) rectangle (13,14)
  (14,13) rectangle (26,14)
  (9,12) rectangle (26,13)
  (1,11) rectangle (3,12)
  (8,11) rectangle (26,12)
  (1,10) rectangle (4,11)
  (7,10) rectangle (26,11)
  (1,9) rectangle (4,10)
  (7,9) rectangle (27,10)
  (1,8) rectangle (4,9)
  (7,8) rectangle (16,9)
  (17,8) rectangle (23,9)
  (24,8) rectangle (27,9)
  (2,7) rectangle (5,8)
  (7,7) rectangle (16,8)
  (17,7) rectangle (23,8)
  (24,7) rectangle (27,8)
  (2,6) rectangle (27,7)
  (3,5) rectangle (26,6)
  (5,4) rectangle (25,5)
  (9,3) rectangle (23,4)
  (10,2) rectangle (23,3)
;
\end{scope}
\node[font=\small] at (82,-3) {Retrieved cat};
\draw[->,thick] (59,16) -- (64,16);
\end{tikzpicture}
\caption{Illustrative stored memories and retrieval from a corrupted input. The lower row shows a noisy cat pattern and the intended recovered memory.}
\label{fig:animalmemories}
\end{figure}

\newpage

\bibliographystyle{plainnat}
\bibliography{netsGames}
\end{document}